\documentclass[runningheads]{llncs}

\usepackage{eccv}

\usepackage{eccvabbrv}

\usepackage{graphicx}
\usepackage{booktabs}

\usepackage[accsupp]{axessibility}  

\usepackage{hyperref}

\usepackage{orcidlink}

\usepackage{multirow}

\begin{document}

\title{Towards Multi-View Consistent Style Transfer with One-Step Diffusion via Vision Conditioning} 

\titlerunning{OSDiffST}

\author{Yushen Zuo \and
Jun Xiao\thanks{Corresponding author} \and 
Kin-Chung Chan \and 
Rongkang Dong\and 
Cuixin Yang \and 
Zongqi He\and 
Hao Xie \and 
Kin-Man Lam
}

\authorrunning{Yushen Zuo et al.}

\institute{The Hong Kong Polytechnic University\\
\email{\{yushen.zuo, kin.man.lam\}@polyu.edu.hk}\\ 
\email{\{jun.xiao, alfred.chen, rongkang97.dong\}@connect.polyu.hk \\ \{cuixin.yang, plume.he, carry-h.xie\}@connect.polyu.hk}
}

\maketitle

\begin{abstract}

The stylization of 3D scenes is an increasingly attractive topic in 3D vision. Although image style transfer has been extensively researched with promising results, directly applying 2D style transfer methods to 3D scenes often fails to preserve the structural and multi-view properties of 3D environments, resulting in unpleasant distortions in images from different viewpoints. To address these issues, we leverage the remarkable generative prior of diffusion-based models and propose a novel style transfer method, OSDiffST, based on a pre-trained one-step diffusion model (i.e., SD-Turbo) for rendering diverse styles in multi-view images of 3D scenes. To efficiently adapt the pre-trained model for multi-view style transfer on small datasets, we introduce a vision condition module to extract style information from the reference style image to serve as conditional input for the diffusion model and employ LoRA in diffusion model for adaptation. Additionally, we consider color distribution alignment and structural similarity between the stylized and content images using two specific loss functions. As a result, our method effectively preserves the structural information and multi-view consistency in stylized images without any 3D information. Experiments show that our method surpasses other promising style transfer methods in synthesizing various styles for multi-view images of 3D scenes. Stylized images from different viewpoints generated by our method achieve superior visual quality, with better structural integrity and less distortion. The source code is available at \url{https://github.com/YushenZuo/OSDiffST}.

\keywords{Multi-view Style Transfer \and One-step Diffusion Models}

\end{abstract}

\section{Introduction}
\label{sec:intro}


Image style transfer is a compelling research area in computer vision that aims to render an image with artistic features derived from a style reference while preserving its original content. Neural style transfer models using deep convolutional neural networks (CNNs) have shown impressive results in synthesizing 2D images with diverse artistic styles \cite{chen2016fast,gatys2016image,gu2018arbitrary,luan2017deep,johnson2016perceptual,wu2020efanet}. In the past years, researchers have attempted to extend these 2D style transfer methods to 3D scenes based on multi-view images, aiming to render the artistic stylized version. However, these 2D methods often struggle to preserve the structural information of objects and the multi-view consistency of 3D scenes, leading to undesirable distortions. Consequently, customizing 3D scenes with desired artistic styles based on multi-view images remains an open and challenging problem.

Recently, large-scale diffusion-based text-to-image models have demonstrated remarkable performance in image synthesis and content creation, enabling users to produce diverse image content in various styles based on given text prompts. By leveraging the advanced generative capabilities of these models, many researchers have explored methods for 3D artistic stylization of scenes using various 3D representations. Yoo \etal \cite{yoo2024plausible} introduced the As-Plausible-as-Possible (APAP) mesh deformation technique, which utilizes 2D diffusion priors to preserve mesh plausibility under user-controlled deformation, resulting in significant improvements. Text2tex \cite{chen2023text2tex} incorporates inpainting techniques into a pre-trained depth-aware image diffusion model to progressively synthesize high-quality textures from multiple viewpoints. Zhuang \etal \cite{zhuang2024tip} proposed a text-to-3D scene editing model that distills prior knowledge from text-to-image diffusion models and manipulates the content and style of input images based on text and image prompts. TexFusion \cite{cao2023texfusion} employs regular diffusion model sampling on different 2D rendered views, demonstrating promising results in texture synthesis. Despite the wide use of diffusion models in various 3D editing and generation tasks, few studies focus on stylizing 3D scenes based on their multi-view images using diffusion models. 

Generally, 2D diffusion-based models have two intrinsic limitations when applied to the stylization of 3D scenes: (1) they typically require numerous sampling steps, often up to tens or hundreds, resulting in a slow inference process, and (2) it is challenging for 2D diffusion models to maintain the multi-view consistency of 3D scenes during the rendering process. Therefore, it is nontrivial to apply these 2D diffusion models to the stylization of multi-view images of 3D scenes.

To address these challenges, we propose OSDiffST, a one-step diffusion model for multi-vew consistent style transfer. OSDiffST effectively synthesizes images from different viewpoints with style reference while preserving image content and multi-view consistency. Our method leverages the diffusion prior from a large-scale pre-trained text-to-image diffusion model called SD-Turbo \cite{sauer2023adversarial} as our generative backbone, which efficiently produces diverse images in a single diffusion sampling step. However, fine-tuning such a large-scale pre-trained diffusion model for new tasks typically requires large image datasets, which is computationally expensive and impractical in many real-world scenarios (e.g., style transfer). To overcome this, we employ the LoRA \cite{hu2021lora} technique into our generative backbone to significantly reduce the number of trainable parameters during fine-tuning, enabling efficient model adaptation for multi-view style transfer. Extracting and injecting style information into the model is crucial for the performance of style transfer. Unlike previous methods based on textual inversion techniques \cite{gal2022image}, our method introduces a vision condition module that uses a pre-trained CLIP \cite{radford2021learning} image encoder and a vision-language projector to extract style information from the reference style image and transfer to the conditional input of our generative backbone for generating stylized images from different viewpoints. To further enhance the quality of stylized content and preserve multi-view consistency, we introduce two loss functions in model training that jointly align the color distribution between the stylized image and the reference style image, and improve the structural similarity between the stylized image and the input content image. 

The main contributions of this paper are summarized as follows:
\begin{enumerate}
    \item We focus on the stylization of multi-view images in 3D scenes and propose OSDiffST, a novel style transfer method based on a one-step diffusion model that effectively preserves the structural information and multi-view consistency of images from different viewpoints.
    
    

    \item To rapidly adapt the pre-trained diffusion model for style transfer, we incorporate LoRA adapters into the pre-trained model, significantly reducing the number of trainable parameters in the training stage. We propose a vision condition module for efficient style information extraction and injection.
    

    \item Our method uses two additional loss functions to align color distribution and improve structural similarity. As a result, our approach enhances visual quality and maintains multi-view consistency across images from different viewpoints without requiring any 3D information.
    
    \item Experiments show that our method has superior capability in rendering artistic styles across images from different viewpoints while preserving multi-view consistency. Compared with other promising style transfer methods, our approach produces results with better visual quality and minimal distortion. 
    
\end{enumerate}

\section{Related Work}
\subsection{Image style transfer}

Neural style transfer is a well-established problem that has been extensively studied over the past few years. The goal of this task is to transfer the style from a reference style image to the input content image. Inspired by the remarkable performance achieved by convolutional neural networks in various vision tasks \cite{simonyan2014very,johnson2016perceptual, xiao2023online,xiao2021feature,xiao2024towards,10385419,10635612,chen2019low}, Gatys \etal \cite{gatys2016image} first proposed an optimization-based method to blend the content and style of given images by minimizing a loss function based on the Gram matrix. Subsequently, Johnson \etal \cite{johnson2016perceptual} introduced image transformation networks to achieve near-optimal results directly, bypassing the need for gradient descent. Huang \etal \cite{huang2017arbitrary} introduced Adaptive Instance Normalization (AdaIN), which performs style transfer by applying the mean and standard deviation of style image features to the normalized content image features. Further, Li \etal \cite{li2017universal} proposed the Whitening and Coloring Transform (WCT) method, which transforms content features to match the statistics of style features.  These methods rely on the pre-trained VGG network \cite{simonyan2014very} for feature extraction and style injection. However, directly applying these 2D style transfer methods to images from multiple viewpoints often fails to preserve structural information and multi-view consistency.

\subsection{3D Style Transfer}

3D style transfer aims to render stylized images from different viewpoints of 3D scenes while preserving structural similarity and multi-view consistency, which is more challenging compared to 2D scenarios. To represent real-world 3D scenes, point clouds, triangle meshes, and radiance fields are widely used in many applications. The methods in \cite{cao2020psnet,bae2023point,mu20223d} first transfer the 3D point clouds into the reference style and then render the corresponding 2D images from different viewpoints. Instead of using point clouds, StyleMesh \cite{hollein2022stylemesh} performs depth-aware patchwise stylization on mesh reconstructions of indoor scenes and achieves promising results. Jin \etal \cite{jin2022language} utilize text prompts as semantic priors to manipulate 3D indoor scenes based on the input 3D mesh. For methods utilizing radiance fields, various NeRF backbones \cite{mildenhall2020nerf,barron2021mip,irshad2023neo, chen2022tensorf,muller2022instant} are used in the stylization of 3D scenes \cite{chiang2022stylizing, chen2022tensorf, huang2022stylizednerf, chen2024upst, pang2023locally, fan2022unified}. However, these methods require images captured from dense views to reconstruct high-quality radiance fields. Although StyleNeRF proposes a zero-shot method, it still struggles with large view shifts. MuVieCAST \cite{ibrahimli2024muviecast} combines 2D style transfer method with 3D geometry learning module for multi-view consistent style transfer in 3D scenes, which requires 3D information (e.g., depth). In contrast to existing methods, our proposed approach leverages geometric properties derived from multi-view images to preserve the structural and multi-view consistency of outputs without using 3D information.

\subsection{Conditional Diffusion-based Generative Models}

Conditional diffusion-based generative models have made significant progress in image manipulation and editing. Unlike conventional editing methods which involve altering specific parts of an image, conditional diffusion-based generative models focus on creating new image content entirely from scratch, guided by specific conditions. GLIDE \cite{nichol2021glide} is a pioneering work that directly generates images in spatial space guided by additional text descriptions. Imagen \cite{saharia2022photorealistic} employs a cascade diffusion framework to progressively generate high-resolution images. Instead of generating in pixel space, Stable Diffusion (SD) \cite{rombach2022high}, VQ-Diffusion \cite{gu2018arbitrary}, and DALL-E-2 \cite{ramesh2022hierarchical} perform the diffusion process in latent space, significantly accelerating the generation speed while maintaining high-quality results. ControlNet-based models \cite{zhang2023adding,zhao2024uni,qin2023unicontrol} provide another framework that incorporates various input types, such as depth maps, normal maps, canny edges, poses, text, and sketches as conditions for image generation. These diffusion-based models require numerous sampling steps for generation, resulting in a slow inference process. Recently, Axel \etal introduced SD-Turbo \cite{sauer2023adversarial}, which utilizes a novel adversarial training approach to enable high quality image generation by stable diffusion with one step sampling. However, directly applying these models for the stylization of 3D scenes typically requires a large dataset for fine-tuning, which is inefficient.

\section{Methodology}
\subsection{Overview of the Proposed Pipeline}
\begin{figure}[tb]
  \centering
  \includegraphics[width=12.2cm]{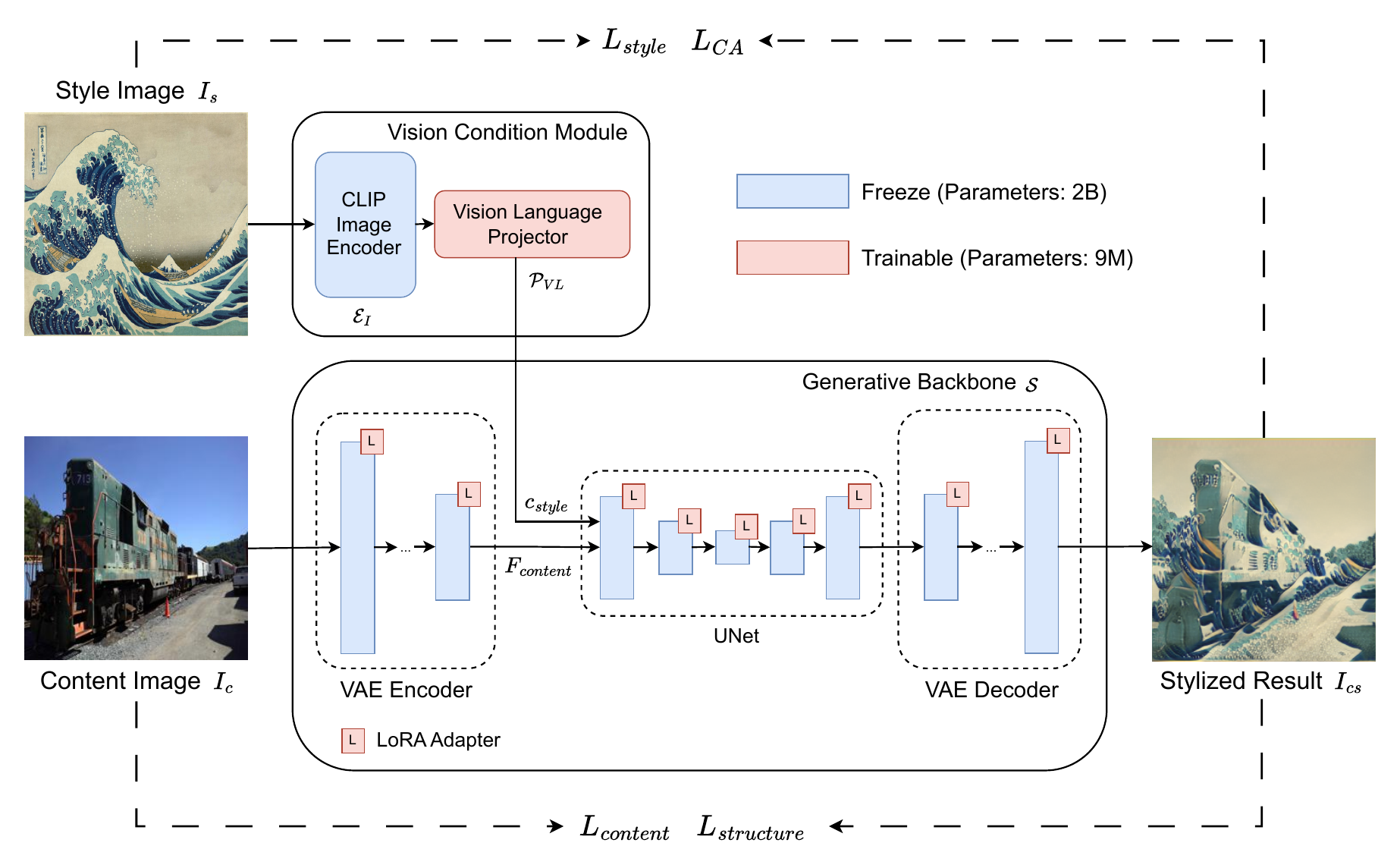}
  \caption{Overview of OSDiffST. The OSDiffST framework consists of two main parts: (1) Vision Condition Module, which includes a pre-trained CLIP Image Encoder $\mathcal{E}_{I}$ and a Vision Language Projector $\mathcal{P}_{\text{VL}}$. This module is used to extract the conditional embedding $c_{\text{style}}$ from the style image $I_{\text{s}}$. (2) Generative Backbone $\mathcal{S}$, which includes a pre-trained one-step stable diffusion model and LoRA adapters. This module is responsible for generating the stylized image $I_{\text{cs}}$ from the input content image $I_{\text{c}}$ and the conditional embedding $c_{\text{style}}$.}
  \label{fig:pipeline}
\end{figure}

Our proposed style transfer model, OSDiffST, effectively preserves the structural information and multi-view consistency properties of multi-view images of 3D scenes. The overall pipeline of OSDiffST is shown in Fig. \ref{fig:pipeline}, which consists of two main components: the generative backbone $\mathcal{S}$ and the vision condition module. For the generative backbone, our method leverages the generative prior from a large-scale pre-trained one-step text-to-image diffusion model (i.e., SD-Turbo). 

To efficiently adapt our proposed model for style transfer on small datasets, our method incorporates LoRA adapters to fine-tune the pre-trained diffusion model in the generative backbone $\mathcal{S}$ with significantly fewer trainable parameters. The vision condition module aims to extract style information from the reference style image $I_{\text{s}}$ by a pre-trained CLIP image encoder \cite{schuhmann2022laionb} $\mathcal{E}_{I}$ and transform the extracted style information to style conditional embedding $c_{\text{style}}$ by our proposed vision language projector $\mathcal{P}_{\text{VL}}$. For a given content image $I_{\text{c}}$, the VAE encoder projects it into latent space, resulting in the latent feature $F_{\text{content}}$. The latent feature $F_{\text{content}}$ and the style conditional embedding $c_{\text{style}}$ are then used as inputs to the UNet, which performs one-step diffusion sampling to obtain the stylized latent embedding. The stylized image is obtained from the stylized latent embedding using the VAE decoder. The whole process can be expressed as $I_{\text{cs}} = \mathcal{S}(I_{\text{c}}, \mathcal{P}_{\text{VL}}(\mathcal{E}_{I}(I_{\text{s}})))$.

In the rest of this section, we will elaborate on each module, including the diffusion-based generative backbone and the vision condition module. Finally, we will describe the loss function used in the training process, which is substantial for the stylization of 3D scenes. 

\subsection{Diffusion-based Generative Backbone}

Text-to-image diffusion models have demonstrated remarkable generative capabilities in image generation and content synthesis. Inspired by this, our proposed method leverages the generative priors from SD-Turbo (v2.1), a large-scale pre-trained text-to-image diffusion model with one-step inference. However, with approximately 1.3 billion parameters, directly adapting this model to other tasks (e.g., style transfer) on small datasets is nontrivial. To address this challenge, we employ LoRA adapter in all convolutional and linear layers in SD-Turbo (v2.1). During the training stage, the parameters of the pre-trained diffusion model are kept fixed, while the parameters of the LoRA adapters are set to be trainable. This allows our method to efficiently encode and fuse the style embedding from the reference style image and content embeddings from the content image, while utilizing the diffusion model's strong generative capabilities for style transfer. Consequently, our method achieves rapid adaptation of the large-scale diffusion model for style transfer on small datasets.

\subsection{Vision Condition Module}

The vision condition module is responsible for extracting style embeddings from style images. It consists of two components: a pre-trained CLIP image encoder and the proposed vision-language projector. CLIP \cite{radford2021learning} has demonstrated a superior ability to learn visual representations and has impressive zero-shot capabilities. Specifically, the CLIP image encoder generates 256 local tokens, which are extracted from the final layer of the transformer block. We combine these local tokens with the global semantic token to form the image embedding. However, the obtained image embedding is unsuitable for direct injection into the generative backbone, as it is not aligned with the corresponding text embedding. Additionally, the dimensionality of the image embedding does not match the requirements for the conditional embedding used by the generative backbone. To address these challenges, we propose a vision language projector $\mathcal{P}_{\text{VL}}$ to transform the image embedding into the corresponding text embedding for alignment. In practice, we use a small MLP network as the vision language projector, which is trainable during the training stage.

\subsection{Loss Function}
As revealed by previous studies \cite{gatys2016image,johnson2016perceptual,cao2020psnet}, the loss functions used to train the network are crucial for the performance and visual quality of translated images. In this paper, our goal is to perform style transfer over multi-view images while preserving image content, geometric structures, and multi-view consistent properties. To achieve this goal, we propose a hybrid loss function that consists of content and style loss, image structural loss, and color alignment loss.


\subsubsection{Content and Style Loss}


Following the studies \cite{gatys2016image,johnson2016perceptual}, the feature maps extracted from the intermediate layers of the pre-trained VGG network contain rich semantic information. Therefore, we define the content loss $L_{\text{content}}$ and style loss $L_{\text{style}}$ used in our method. The content loss is expressed as follows:
\begin{equation}
    L_{content} = \sum_{\ell} L_{1}^{\text{s}}(\phi_{\ell}(I_{\text{cs}}), \phi_{\ell}(I_{\text{c}})),\quad \ell\in \{3, 4, 5\}, 
\end{equation}
where $\phi_{\ell}(\cdot)$ denotes the features extracted from the $\ell$-th layer followed by the ReLU function in the ImageNet \cite{deng2009imagenet} pre-trained VGG-19 network and $L_{1}^{\text{s}}$ represents the smooth $L_{1}$ loss which is defined as follows:
\begin{equation}
    L_{1}^{\text{s}}(x) = \begin{cases} 0.5x^{2},\quad \text{if}\ \vert x\vert < 1; \\
    \vert x\vert -0.5,\quad \text{others}.
    \end{cases}
\end{equation}
We adopt the smooth $L_{1}$ loss function because it is differentiable in small values and stabilizes the training process. This content loss function aims to preserve the image content during the process of style transfer. 

Following the previous studies \cite{gatys2016image}, we adopt the Gram matrix between the feature maps to measure the style distance. The style loss is defined as follows:
\begin{equation}
    L_{\text{style}} = \sum_{l=1}^{5} \frac{1}{N_{\ell}} \left \Arrowvert G^{\ell}_{I_{\text{cs}}} - G^{\ell}_{I_{\text{c}}} \right \Arrowvert_{2}^{2},
\end{equation}
where $G^{\ell}_{I_{\text{cs}}}$ and $G^{\ell}_{\text{c}}$ are Gram matrices of $\ell$-th layer feature extracted from the style image $I_{\text{cs}}$ and the content image $I_{\text{c}}$ in the pre-trained VGG-19 network.



\subsubsection{Image Structure Loss}

To encourage content preservation and multi-view consistency, we further propose a structural loss denoted by $L_{\text{structure}}$, which enforces the model to preserve the geometric characteristics of images. In particular, we extract various geometric features from the input content images $I_{\text{c}}$ and the stylized image $I_{\text{cs}}$ using Sobel operator, Laplacian operator, and Canny edge detector. Based on the extracted features, the image structural loss function is calculated as follows:
\begin{equation}
    L_{structure} = \sum_{i\in \{S, L, C\}} L_{1}^{\text{s}}(O_{i}(I_{cs}), O_{i}(I_{c})),
\end{equation}
where $O=\{\text{S}:\text{Sobel operator},\ \text{L}:\text{Laplacian operator},\ \text{C}: \text{Canny operator}\}$ denotes a set of image processing operators.

\subsubsection{Color Alignment Loss}
Color is a significant factor affecting the rendered style and perceptual quality of the output image. To ensure the color style of the translated image closely matches the reference style images, we propose using a color alignment loss $L_{\text{CA}}$ based on HistoGAN \cite{afifi2021histogan}, which is defined as follows:
\begin{equation}
    L_{CA} = \frac{1}{\sqrt{2}} \left \Vert H(I_{cs})^{1/2} - H(I_{s})^{1/2} \right \Vert_{2},
\end{equation}
where $H(\cdot)$ is the color histogram of an image. The color alignment loss aims to encourage the stylized image $I_{cs}$ to follow the color palettes of the reference style image $I_{s}$.


The overall loss function in our framework is the weighted sum of the above loss terms as follows:
\begin{equation}
    L = \lambda_{content}L_{content} + \lambda_{style}L_{style} + \lambda_{structure}L_{structure} +\lambda_{CA}L_{CA},
\end{equation}
where $\lambda_{content}, L_{content}, \lambda_{style}, \lambda_{structure}$, and $\lambda_{CA}$ are hyper-parameters to balance different loss functions. Therefore, our method can generate translated images with various styles by adjusting these hyper-parameters.

\section{Experiments}

In this section, we describe the details of our experiments and analyze the results by comparing our proposed method with other promising CNN-based and diffusion-based style transfer methods. Finally, we conduct a comprehensive ablation study to analyze the effects of different components of our method in multi-view consistent style transfer.

\subsection{Implementation Details}

In our experiments, we adopt the public Tanks and Templates dataset \cite{Knapitsch2017}, which contains 14 scenes, each with 300 to 500 images captured from different viewpoints. Our proposed method OSDiffST contains a large-scale pre-trained text-to-image diffusion model (i.e., SD-Turbo \cite{sauer2023adversarial}) with additional LoRA adapters, a pre-trained CLIP image encoder \cite{schuhmann2022laionb}, and a vision language projector. In the training stage, we freeze the model parameters of SD-Turbo and the CLIP image encoder, training only the LoRA adapters and the vision-language projector. Consequently, the trainable parameters of our method are much fewer than the full fine-tuning setting (9M v.s. 2B), leading to efficient model adaptation over small datasets for style transfer. We use the Adam optimizer \cite{KingBa15} to adaptively update the model parameters during training with an initial learning rate of $1\times 10^{-3}$ and a batch size of $3$. Images are resized to $256\times 256$ in both training and testing stages. As our method adopts a hybrid loss function, the hyperparameters used in the loss function are substantial to the content and visual quality of the outputs. In implementation, we set $\lambda_{\text{content}}=1\times 10^{3}$, $\lambda_{\text{style}}=1\times 10^{8}$, $\lambda_{\text{structure}}=2\times 10^{4}$, and $\lambda_{\text{CA}}=1\times 10^{4}$. Our method is implemented with the PyTorch framework and a single Nvidia RTX 3090 GPU. In particular, OSDiffST is trained on each individual scene for style transfer with a single reference style image, which takes approximately 30 minutes to complete training.


\subsection{Experiment Results}

In our experiments, we compared our proposed method with other promising style transfer methods, including AdaIN \cite{huang2017arbitrary}, StyleFormer \cite{wu2021styleformer}, MuVieCAST\footnote{We use AdaIN as the TransferNet in MuVieCAST in all our experiments.} \cite{ibrahimli2024muviecast}, and InST\footnote{We use the recommended setting in InST with strength = 0.7 in all our experiments.} \cite{zhang2023inversion}. For quantitative comparison, we adopt two metrics for evaluation: 1) Color Histogram Distance (CHD) \cite{afifi2021histogan} and 2) DINO Structure Distance (DSD) \cite{tumanyan2022splicing}. CHD calculates the Hellinger distance of the color histogram between the stylized image and the reference style image. A lower score indicates better similarity in color distribution between the stylized image and the reference style image. DSD measures the structural similarity between the stylized images and the given content image, with all scores multiplied by 100. A lower score indicates a more accurate preservation of the content structure in the stylized image. For qualitative comparison, we compare different methods via visual results and conduct a user study for evaluation. Specifically, in the user study, we invite participants to score the stylized images on two aspects: content preservation and stylization. Participants are asked to score stylized images from 1 to 5 in each aspect: 1 - very bad, 2 - bad, 3 - medium, 4 - good, and 5 - very good. We calculate the average score for evaluation and comparison.

\begin{figure}[tb!]
  \centering
  \includegraphics[width=12.2cm]{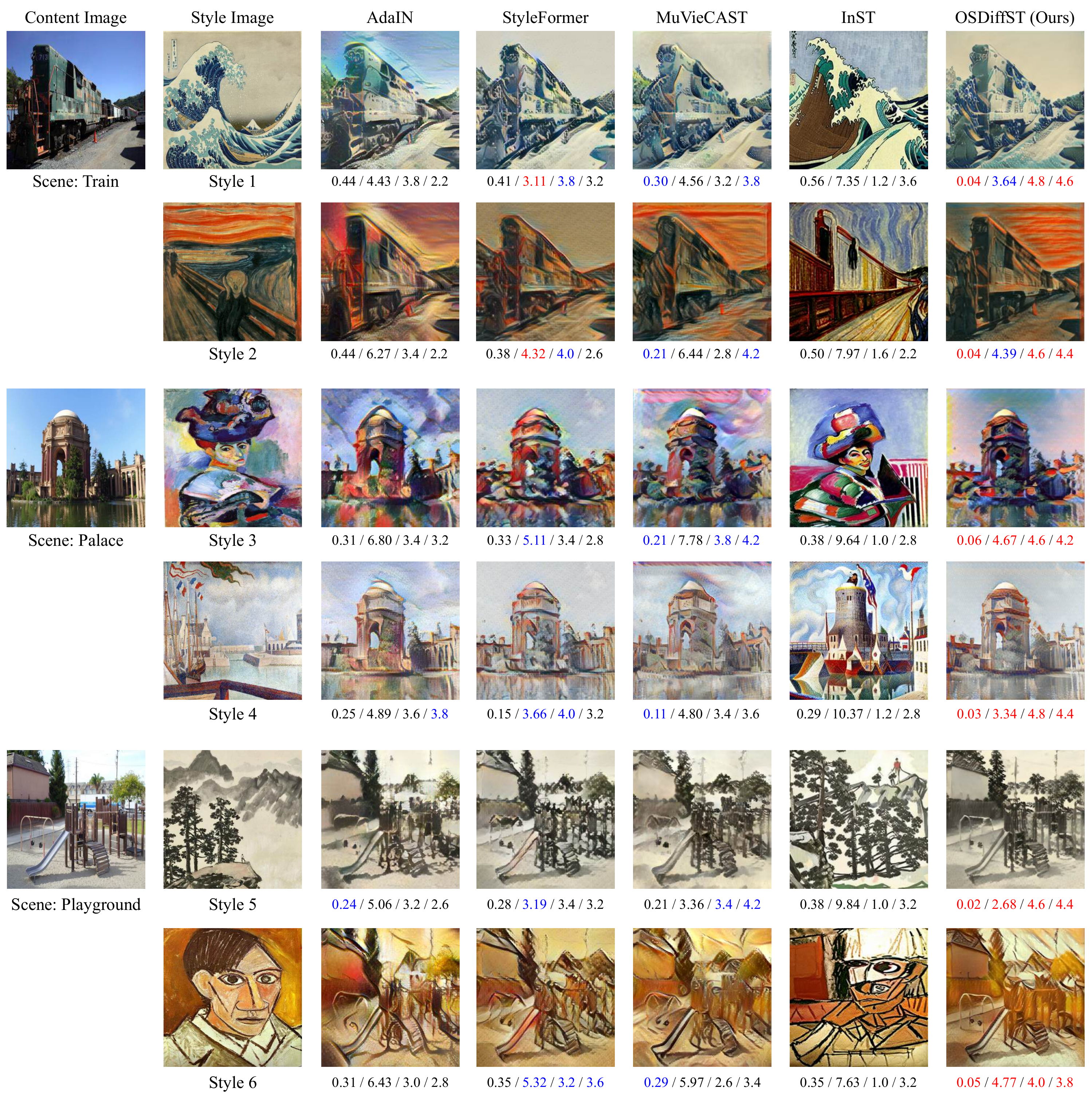}
  \caption{Experimental results of different style transfer methods. We use two objective metrics and two subjective metrics to evaluate stylized images. `CHD' and `DSD' are used as objective metrics. For subjective metrics, we calculate the average score of `Content Preservation' and `Stylization' from the user study, which denoted as `Avg. Content Preservation' and `Avg. Stylization', respectively. Metrics are shown in `CHD' / `DSD' / `Avg. Content Preservation' / `Avg. Stylization'. The best and second-best results are highlighted in red and blue, respectively.}
  \label{fig:exp_compare}
\end{figure}

We conducted experiments on three scenes from the Tanks and Temples dataset: `Train', `Playground', and `Palace', using six styles. The visual comparison and corresponding metrics are shown in Fig. \ref{fig:exp_compare}. Quantitative metrics show that stylized images from our method have better structural similarity to the input content image and a better color histogram match to the style image. User study results align with the quantitative results, demonstrating that our method outperforms other methods in content preservation and stylization in style transfer. It is worth noting that our method is trained on a much smaller dataset (300 $\sim$ 500 images) compared to CNN-based methods \cite{huang2017arbitrary, wu2021styleformer}, which use the MSCOCO dataset (5000 images) \cite{lin2014microsoft} for training. In addition, we conducted a quantitative evaluation of all images in each scene. Specifically, we transferred all images in each scene to the target style using methods shown in Fig. \ref{fig:exp_compare} and calculated the average CHD and DSD metrics. Experiment results are shown in Table \ref{tab:exp_compare}. Our method achieves the best performance in most cases in terms of both metrics. 

\begin{table}[]
\caption{Quantitative results in each scene. Metrics are shown in `CHD' / `DSD'. The best and second-best results are highlighted in red and blue, respectively.}
\label{tab:exp_compare}
\centering
\begin{tabular}{c|cc|cc|cc}
\hline
\multirow{2}{*}{Method} & \multicolumn{2}{c|}{Train}                     & \multicolumn{2}{c|}{Palace}                    & \multicolumn{2}{c}{Playground}                 \\ \cline{2-7} 
                        & \multicolumn{1}{c|}{Style 1}     & Style 2     & \multicolumn{1}{c|}{Style 3}     & Style 4     & \multicolumn{1}{c|}{Style 5}     & Style 6     \\ \hline
AdaIN \cite{huang2017arbitrary}                  & \multicolumn{1}{c|}{0.43/\textcolor{blue}{3.44}} & 0.44/4.22 & \multicolumn{1}{c|}{0.28/6.17} & 0.24/4.69 & \multicolumn{1}{c|}{0.21/4.34} & 0.32/7.50 \\ \hline
StyleFormer \cite{wu2021styleformer}            & \multicolumn{1}{c|}{0.42/\textcolor{red}{3.35}} & 0.35/\textcolor{red}{3.75} & \multicolumn{1}{c|}{0.28/\textcolor{blue}{5.07}} & 0.16/\textcolor{blue}{3.96} & \multicolumn{1}{c|}{0.28/3.78} & 0.37/\textcolor{blue}{6.54} \\ \hline
MuVieCAST \cite{ibrahimli2024muviecast}              & \multicolumn{1}{c|}{ \textcolor{blue}{0.32}/4.02} & \textcolor{blue}{0.21}/4.64 & \multicolumn{1}{c|}{\textcolor{blue}{0.21}/6.32} & \textcolor{blue}{0.13}/5.06 & \multicolumn{1}{c|}{\textcolor{blue}{0.20}/\textcolor{blue}{4.06}} & \textcolor{blue}{0.32}/6.99 \\ \hline
InST \cite{zhang2023inversion}                   & \multicolumn{1}{c|}{0.46/6.80} & 0.40/6.56 & \multicolumn{1}{c|}{0.39/7.44} & 0.30/8.81 & \multicolumn{1}{c|}{0.32/9.79} & 0.42/8.58 \\ \hline
OSDiffST (Ours)                    & \multicolumn{1}{c|}{ \textcolor{red}{0.04}/3.81} &  \textcolor{red}{0.05}/\textcolor{blue}{3.79} & \multicolumn{1}{c|}{\textcolor{red}{0.07}/\textcolor{red}{4.69}} & \textcolor{red}{0.03}/\textcolor{red}{3.65} & \multicolumn{1}{c|}{\textcolor{red}{0.02}/\textcolor{red}{2.61}} & \textcolor{red}{0.05}/\textcolor{red}{5.88} \\ \hline
\end{tabular}
\end{table}

To show the superiority of one-step diffusion, we also tested the running time of each method at the spatial resolutions of 128$\times$128 and 256$\times$256 using a single RTX 3090. The inference time comparison is shown in Table \ref{tab:runningtime}. These results show that with one-step diffusion sampling, our method is comparable to CNN-based methods in inference time and more efficient than InST, a multi-step diffusion-based method.

\begin{table}[tb]
  \caption{Inference time (in seconds) of different methods at spatial resolutions of 128$\times$128 and 256$\times$256.}
  \label{tab:runningtime}
  \centering
  \begin{tabular}{@{}c|c|c|c|c|c@{}}
    \hline
    Resolution & AdaIN \cite{huang2017arbitrary} & StyleFormer \cite{wu2021styleformer} & MuVieCAST \cite{ibrahimli2024muviecast} & InST \cite{zhang2023inversion} & OSDiffST (Ours) \\
    \hline
    $128\times128$  & 0.001 & 0.003 & 0.001 & 4.0 & 0.028 \\
    \hline
    $256\times256$  & 0.003 & 0.008 & 0.003 & 4.2 & 0.063 \\
  \hline
  \end{tabular}
\end{table}

\begin{figure}[tb!]
  \centering
  \includegraphics[width=12.2cm]{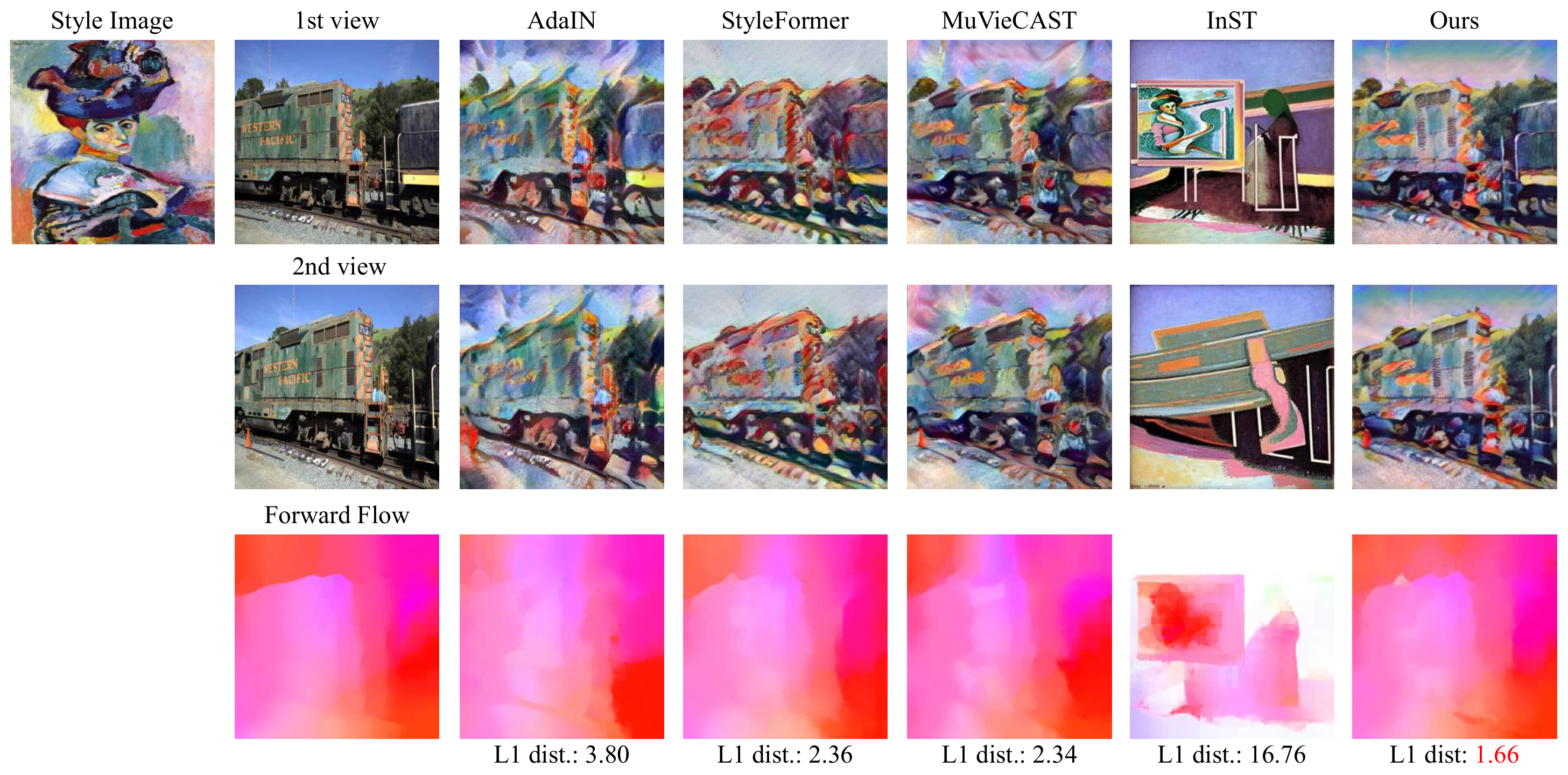}
  \caption{Stylized results of different methods in a multi-view scenario. `L1 dist.' denotes the L1 distance between the forward flow from stylized images and content images.}
  \label{fig:mvc}
\end{figure}

We further compare our method with baseline methods in a multi-view scenario. Specifically, we compare the visual results of stylized images in two consecutive views randomly selected from the `Train' scene to assess the consistency of the primary structure of objects in the content image. We first estimate the forward flow from the first view to the second view via GMFlow \cite{xu2022gmflow}. Then, we evaluate multi-view consistency by calculating the L1 distance between the forward flow from images. As shown in Fig. \ref{fig:mvc}, the forward flow of stylized images from our method has the lowest L1 distance to the forward flow from content images. This experimental result shows that our method achieves the best consistency of content in stylized images under a multi-view scenario.

\subsection{Ablation study}
In this part, we explore our proposed method in four aspects: (1) Color Alignment Loss, (2) Vision condition module, (3) Generalization, and (4) Multi-view Consistency.

\subsubsection{Color Alignment Loss}
\begin{figure}[tb]
  \centering
  \includegraphics[width=12.2cm]{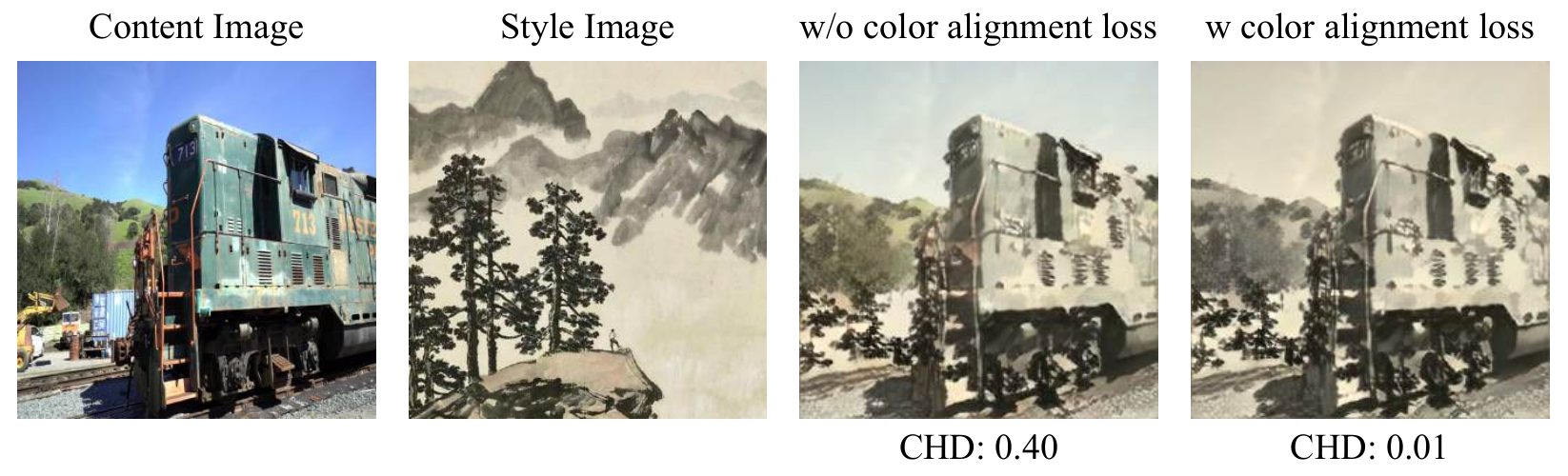}
  \caption{Stylized results from different configurations of the color alignment loss in model training.}
  \label{fig:ablation_1}
\end{figure}

In this section, we conducted an ablation study on color alignment loss $L_{CA}$ to explore its contribution to style transfer. Specifically, we train OSDiffST with or without color alignment loss in this ablation study. The visual results and quantitative metric comparison are shown in Fig. \ref{fig:ablation_1}. 

When training with color alignment loss, the CHD metric of the stylized image greatly decreases, indicating a better color histogram distribution matching between the stylized image and the reference style image. Experimental results reveal that training with the color alignment loss results in stylized images closer to the style image in the color space, achieving better style transfer results.


\subsubsection{Vision condition module}
\begin{figure}[tb]
  \centering
  \includegraphics[width=12.2cm]{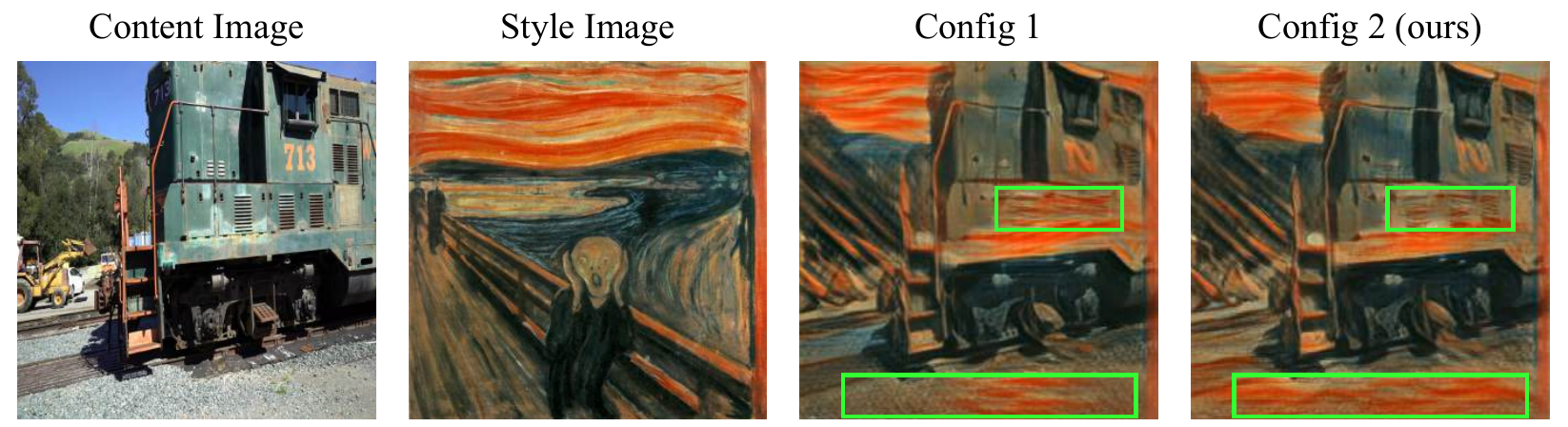}
  \caption{Stylized results from different configurations of generating condition embedding.}
  \label{fig:ablation_2}
\end{figure}

In this section, we explore the vision condition module in our framework. Specifically, we conducted an ablation study by training our network with two configurations of condition embedding: Config 1: randomly initialized learnable embedding, and Config 2: condition embedding from the proposed vision condition module. Under the same experimental setting, the visual comparison of stylized images is shown in Fig. \ref{fig:ablation_2}.

From the visual comparison results, images obtained by the network after training with Config 1 exhibit inconsistent style injection, as highlighted in the green box in Fig. \ref{fig:ablation_2}. For areas with continuous semantics, such as the ground, the stylized image obtained by Config 1 is not rendered uniformly, resulting in inconsistent style transfer and affecting the quality of the output image. In contrast, the stylized image obtained by Config 2 alleviates this problem.

\subsubsection{Generalization}
\begin{figure}[tb]
  \centering
  \includegraphics[width=12.2cm]{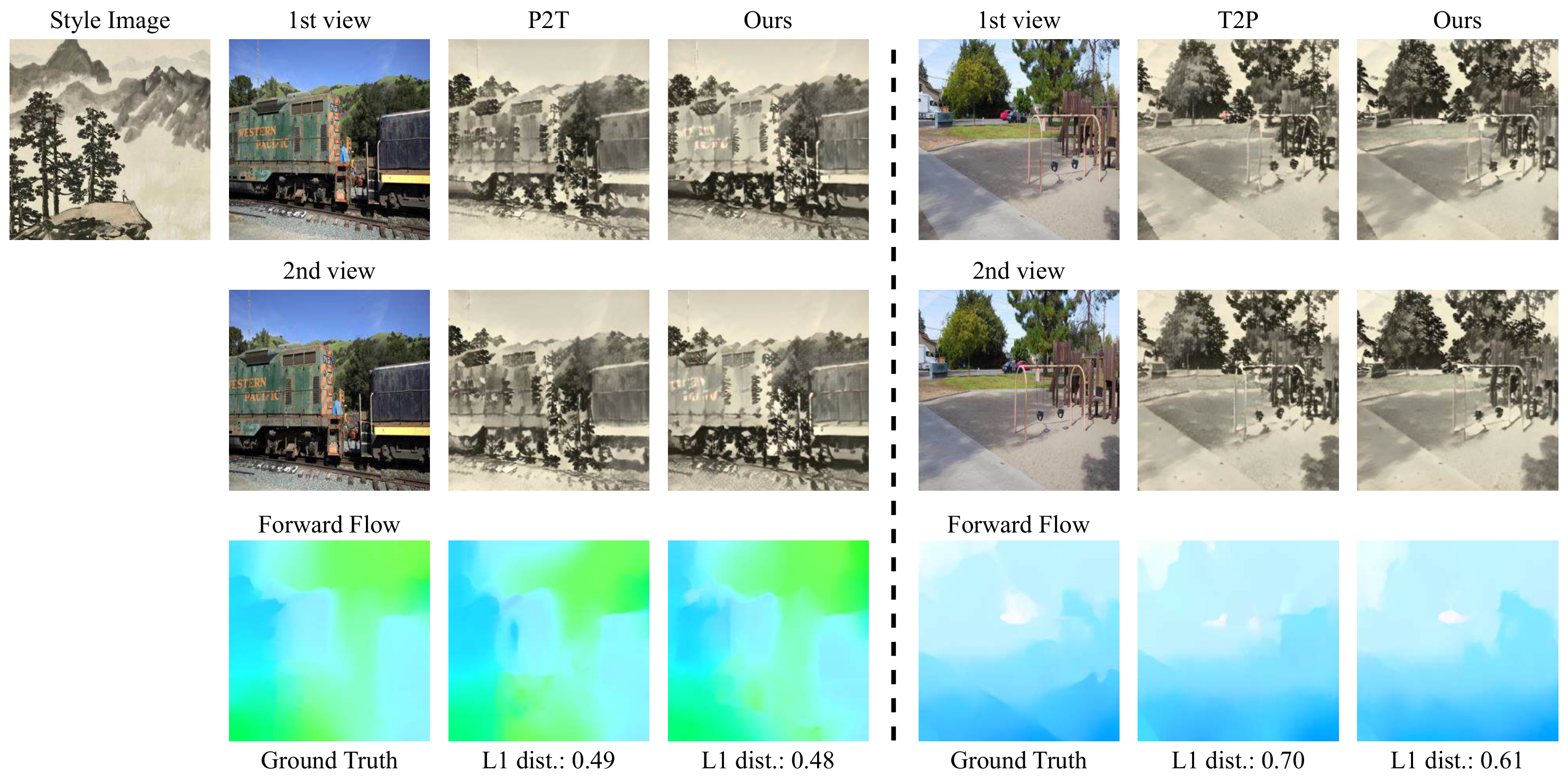}
  \caption{Stylized results and multi-view consistency evaluation from different training strategies.}
  \label{fig:ablation_3}
\end{figure}

In this section, we explore the generalization ability of our proposed method. Specifically, we train OSDiffST under two training configurations: (1) Train on the `Train' scene and test on the `Playground' scene, denoted as `T2P', and (2) Train on the `Playground' scene and test on the `Train' scene, denoted as `P2T'. After training, we compare the stylized images with results from the original setting. Visual comparisons of stylized images and their corresponding forward flow are shown in Fig. \ref{fig:ablation_3}.

Experimental results show that even training with data from different scenes, our method generates results similar to the original setting in terms of the visual quality of stylized images and the L1 distance of forward flow, indicating that they also achieve multi-view consistent style transfer. This experiment demonstrates that our method is not limited to learning to render a single scenario, but has the generalization ability to transfer new scenarios to the target style while maintaining multi-view consistency. This highlights the high applicability of our method.

\subsubsection{Multi-view Consistency}
In this section, we explore the multi-view consistency of our proposed method. Specifically, we analyze the impact of structure loss by comparing stylized results from two loss configurations in training: Config 1: Training without structure loss, and Config 2: Training with structure loss.

\begin{figure}[tb]
  \centering
  \includegraphics[width=9.0cm]{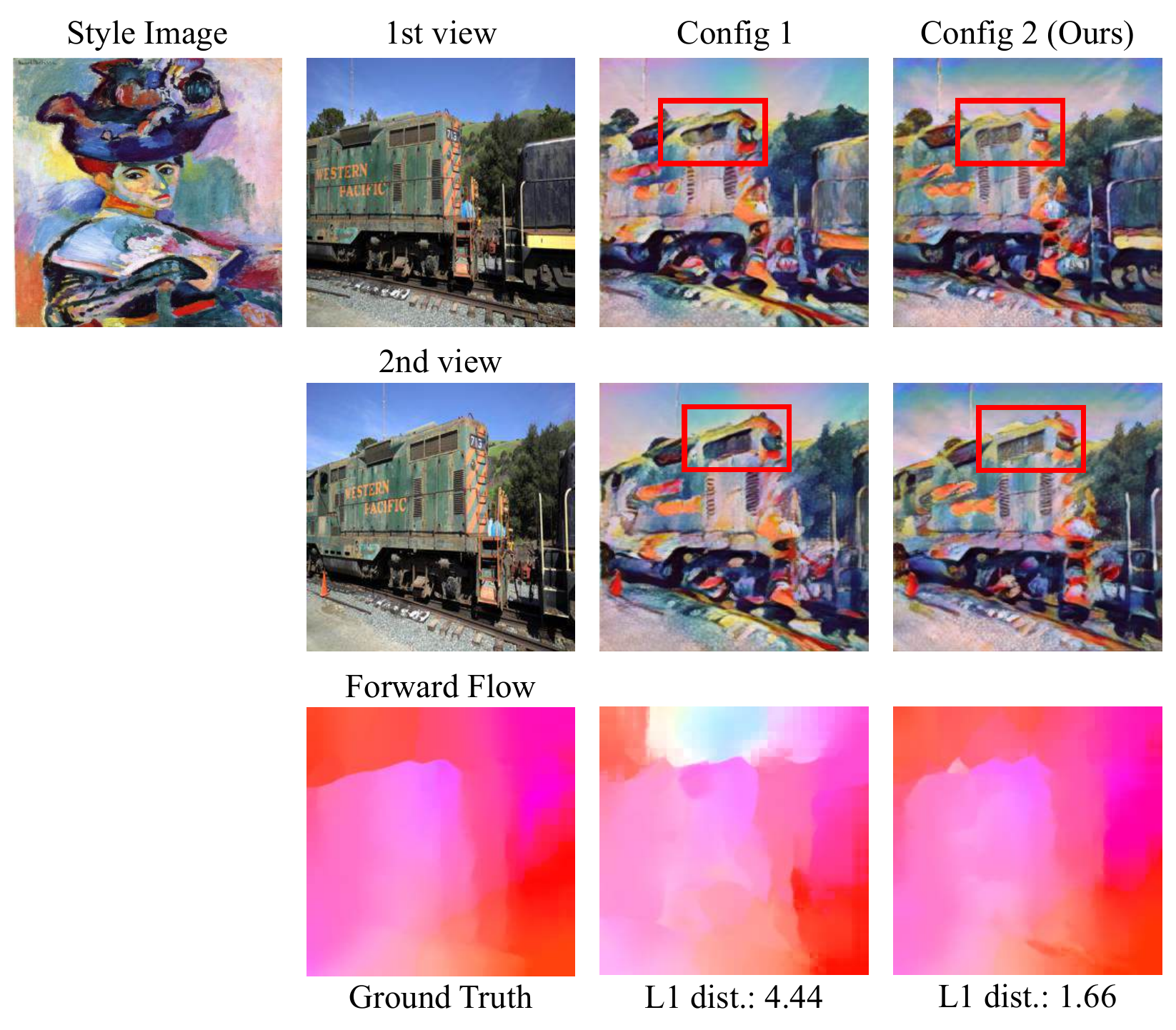}
  \caption{Stylized results from different configurations of structure loss in training.}
  \label{fig:ablation_4}
\end{figure}

As shown in Fig. \ref{fig:ablation_4}, training with structure loss significantly decreases the L1 distance of forward flow. A detailed visual comparison is highlighted within the red box. When training without structure loss, the primary object from the content image undergoes noticeable changes in the stylized image under different viewpoints. The model trained with structure loss alleviates this problem. This experiment demonstrates that structure loss greatly enhances the multi-view consistency of our method.

\section{Conclusion}
In this work, we introduce OSDiffST, a one-step diffusion model designed to transfer images to target styles from various viewpoints while maintaining image content and consistency across views. We incorporate LoRA adapters in pre-trained SD-Turbo to build our generative backbone for generating images in a single diffusion step. We also propose a vision condition module that utilizes pre-trained CLIP image encoder and a vision language projector to extract and inject style information from the reference style image into the generative backbone as its conditional input. We further integrate color alignment loss and image structure loss during model training to enhance style transfer and multi-view consistency. Experimental results demonstrate that our method successfully transfers content image in different views to the target style in one diffusion step while maintaining multi-view consistency.

\subsubsection{Acknowledgements}
This work was supported by the Hong Kong Research Grants Council (RGC) Research Impact Fund (RIF) under Grant R5001-18.

\bibliographystyle{splncs04}
\bibliography{main}
\end{document}